\documentclass[letterpaper, 10 pt, conference]{ieeeconf}  % Comment this line out if you need a4paper

\IEEEoverridecommandlockouts                              % This command is only needed if 
\usepackage{cite}
\usepackage{amsmath,amssymb,amsfonts}
\usepackage{algorithmic}
\usepackage{graphicx}
\usepackage{textcomp}
\usepackage{xcolor}
\usepackage{float}
\usepackage{url}
\usepackage{booktabs}
\usepackage{array}
\usepackage{multirow} 
\usepackage{comment}
\usepackage[caption=false,font=footnotesize]{subfig}
\usepackage{booktabs}
\usepackage{tabularx}
\usepackage{makecell}
\usepackage{adjustbox}

\newcommand{\name}{vAccSOL}

\def\BibTeX{{\rm B\kern-.05em{\sc i\kern-.025em b}\kern-.08em
    T\kern-.1667em\lower.7ex\hbox{E}\kern-.125emX}}

\newif\ifanonym
\anonymfalse     % <-- camera-ready

\ifanonym
  \usepackage{censor}
\fi

\begin{document}

\title{\name{}: Efficient and Transparent AI Vision Offloading for Mobile Robots}

\ifanonym
\author{Anonymous Authors} % or "Authors withheld for double-blind review"
\else
\author{%
Adam Zahir$^{1}$,
Michele Gucciardo$^{2}$,
Falk Selker$^{2}$,
Anastasios Nanos$^{3}$,\\
Kostis Papazafeiropoulos$^{3}$,
Carlos J. Bernardos$^{1}$,
Nicolas Weber$^{2}$,
Roberto Gonzalez$^{2}$%
\thanks{$^{1}$University Carlos III of Madrid (UC3M), Madrid, Spain.
{\tt\small \{name.surname\}@uc3m.es}}%
\thanks{$^{2}$NEC Laboratories Europe, Heidelberg, Germany.
{\tt\small \{name.surname\}@neclab.eu}}%
\thanks{$^{3}$Nubificus Ltd, United Kingdom.
{\tt\small \{name.surname\}@nubis-pc.eu}}%
}
\fi

\maketitle
\thispagestyle{empty}
\pagestyle{empty}

\begin{abstract}
Mobile robots are increasingly deployed for inspection, patrol, and search-and-rescue operations, relying on computer vision for perception, navigation, and autonomous decision-making. However, executing modern vision workloads onboard remains challenging due to limited computational resources and strict energy constraints. Although commercial platforms may include embedded accelerators, these are often tied to proprietary software stacks, leaving user-defined vision tasks to run on resource-constrained companion computers.

To address this limitation, we present \emph{\name{}}, a framework for efficient and transparent execution of AI-based vision workloads across heterogeneous robotic and edge platforms. The proposed solution integrates two complementary components: the SOL neural network compiler and the vAccel execution framework. SOL compiles deep learning models into optimized inference libraries with minimal runtime dependencies, while vAccel provides a lightweight, hardware-agnostic execution layer that transparently dispatches inference either locally on the robot or remotely to nearby edge infrastructure. Together, these components enable hardware-optimized inference and flexible execution placement without requiring modifications to the robot application.

We implement and validate \emph{\name{}} on a real-world testbed using a commercial quadruped robot and twelve deep learning models relevant to mobile robotic operations, spanning image classification, video classification, and semantic segmentation. Compared to a PyTorch compiler baseline, SOL achieves comparable or improved inference performance. With edge offloading, \emph{\name{}} reduces robot-side power consumption by up to 80\% compared to local execution and edge-side power consumption by up to 60\% compared to PyTorch, while increasing the vision pipeline frame rate by up to 24$\times$, thereby extending the operating lifetime of battery-powered robots.
\end{abstract}

% Keywords
%Computer vision, offloading, 5G, robotics, neural network optimization

%%%%%%%%%%%%%%%%%%%%%%%%%%%%%%%%%%%%%%%%%%%%%%%%%%%%%%%%%%%%%%%%%%%%%%%%%%%%%%%%
\section{Introduction}
Mobile robots are increasingly deployed in complex, unstructured, and hazardous environments, including industrial inspection, autonomous patrolling, logistics, and search-and-rescue operations~\cite{mdpi2025quadruped}. These applications demand robust perception and real-time decision-making, making computer vision a core enabling technology. Through continuous processing of visual and multimodal sensor data, vision systems support perception, localization, navigation, and autonomous control across diverse platforms, including wheeled robots, quadrupeds, and aerial drones. 
% Among these, quadruped robots have gained particular attention due to their ability to traverse stairs, rubble, uneven terrain, and confined industrial spaces where wheeled systems are ineffective.

Despite rapid advances in deep learning for visual perception, executing advanced vision models onboard mobile robots remains challenging. Mobile platforms are inherently resource-constrained due to size, payload, and battery capacity~\cite{ichnowski2023fogros2}. At the same time, robots integrate multiple RGB cameras, depth sensors, LiDAR, and thermal imagers, generating high-rate data streams that place substantial strain on embedded processors. Running tasks such as object detection, semantic segmentation, and simultaneous localization and mapping (SLAM) in real time leads to increased energy consumption and reduced operational endurance. As robots become ubiquitous, onboard computational resources must remain relatively small to reduce energy usage and manufacturing cost, yet perception and decision-making systems are becoming increasingly computationally expensive~\cite{chinchali2019network}.

Although commercial robotic platforms may include embedded GPU accelerators (e.g., NVIDIA Jetson modules), these resources are often tightly integrated with proprietary control and navigation stacks, limiting their availability for user-defined vision workloads. Deploying custom inference pipelines on vendor-controlled platforms may require modifying system software, voiding warranties, or competing for resources with safety-critical control loops. In practice, constraints on mobility, power, and cost prevent robots from running modern algorithms at the desired rates~\cite{ichnowski2023fogros2}. As a result, many research and industrial deployments rely on external compute platforms—such as portable mini-PCs or companion devices—that offer unrestricted development access but lack dedicated GPU acceleration. This practical limitation motivates our focus on computation offloading as a primary strategy for enabling advanced vision inference on commercially available mobile robots.

Edge and fog computing address this challenge by deploying computational resources close to the data source or point of operation, such as at wireless access points, local servers, or network edge facilities. These resources can be accessed through wireless technologies including Wi-Fi and cellular networks, enabling flexible deployment across diverse environments~\cite{kehoe2015cloud}. A wide range of computer vision tasks in mobile robotics benefit from offloaded execution, including semantic segmentation for navigation and video classification for surveillance. Executing these workloads on nearby edge servers enables the use of larger and more accurate models without exceeding onboard resource constraints, while maintaining the responsiveness required for safe operation. Moreover, offloading provides energy advantages. Even when onboard accelerators are available, their power consumption (e.g., 7–25W for embedded GPUs~\cite{nvidia2024jetson}) may exceed the energy cost of transmitting data to a nearby edge server and receiving inference results. For battery-powered mobile robots, reducing onboard computation directly extends operational lifetime. Efficient and transparent offloading mechanisms are therefore essential to balance performance and energy efficiency in real-world deployments.

To address these challenges, we propose \emph{\name{}}, a unified framework for accelerating and offloading AI vision workloads on mobile robots. 
\emph{\name{}} integrates two key components: 
(i) the SOL compiler, which generates hardware-optimized inference binaries for heterogeneous accelerators, and 
(ii) vAccel, a lightweight API remoting system that enables inference workloads to execute transparently either locally on the robot or remotely on nearby edge infrastructure without requiring modifications to the robot vision application.
Together, these components enable hardware-agnostic AI acceleration and seamless execution across the robot--edge computing continuum.

The main contributions of this paper are:
\begin{itemize}
    \item We present \emph{\name{}}, an AI acceleration framework that integrates compiler-level neural network optimization with edge computing to enable efficient, scalable, and transparent execution of vision workloads on mobile robots.

    \item We implement and validate \emph{\name{}} on a real-world testbed that integrates a commercial off-the-shelf quadruped robot with local edge infrastructure.

    \item We conduct a comprehensive performance evaluation across twelve deep learning models spanning image classification, video action recognition, and semantic segmentation. Our results show that edge offloading (i) reduces robot-side power consumption by up to 80\% compared to local execution and edge-side power consumption by up to 60\% compared to PyTorch, and (ii) increases the vision pipeline frame rate by up to 24$\times$ for compute-intensive models.
\end{itemize}

The remainder of this paper is organized as follows. 
Section~\ref{sec:SoA} reviews related work. 
Section~\ref{sec:architecture} describes the proposed system architecture. 
Section~\ref{sec:evaluation} presents the experimental setup and evaluation 
methodology. Section~\ref{sec:results} reports and discusses the results. 
Finally, Section~\ref{sec:conclusion} concludes the paper and outlines 
future research directions.
\section{Related work}
\label{sec:SoA}

\begin{figure}[tb]
   \centering    
   \includegraphics[height=0.85\columnwidth]{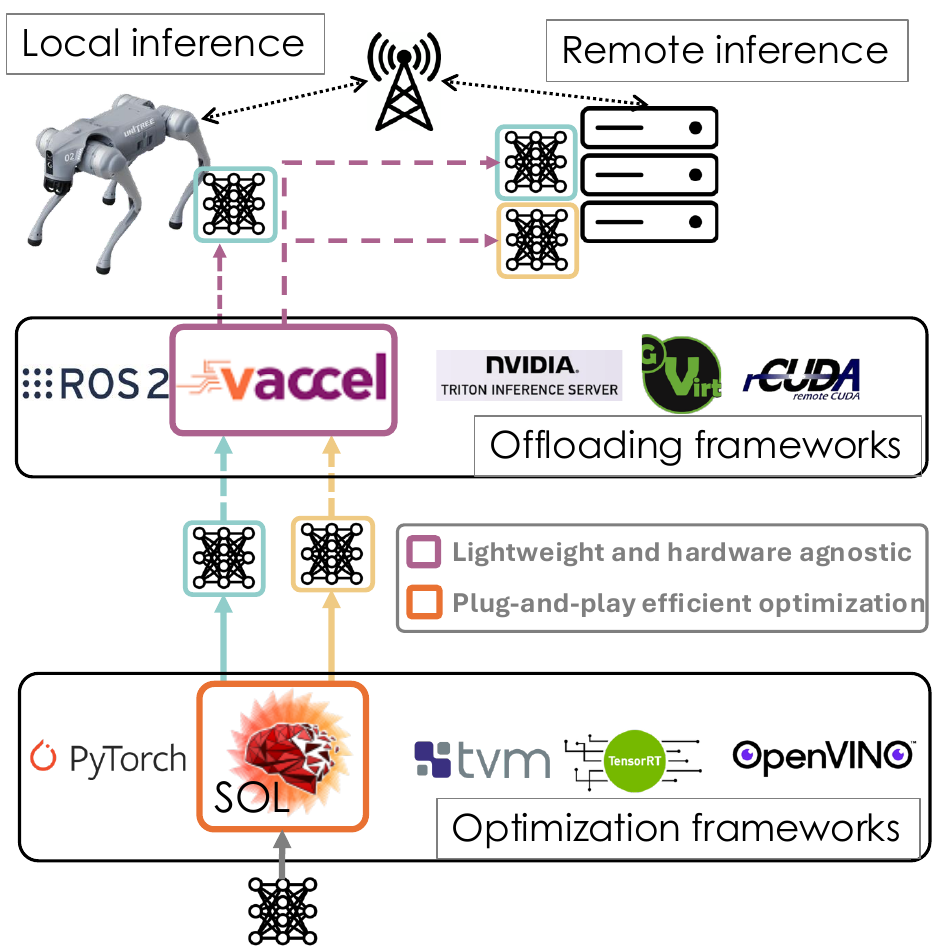}
      \caption{Landscape of optimization and offloading frameworks.}
      \label{fig:landscape}
      \vspace{-2mm}
\end{figure}

This section reviews prior work on computation offloading for mobile robots, neural network optimization for edge deployment, and transparent acceleration frameworks. Figure~\ref{fig:landscape} summarizes the landscape of existing optimization and offloading approaches.

\subsection{Computation Offloading for Mobile Robots}

Edge and fog computing have been widely studied as mechanisms to augment the computational capabilities of resource-constrained robots. By offloading perception workloads to nearby infrastructure, robots can execute larger models while reducing onboard energy consumption. A comprehensive survey is provided in~\cite{afrin2025edge}, covering localization, navigation, and perception across ground robots, drones, and multi-robot systems. Experimental studies demonstrate clear benefits. Edge-assisted object detection reduces latency compared to local or cloud-only execution~\cite{liu2019edge}. Offloading SLAM and visual odometry improves frame rate and reduces onboard resource usage~\cite{sarker2019slam,qingqing2019odometry}, while semantic visual SLAM leverages edge resources to enable real-time scene understanding~\cite{xu2020semantic}. These benefits, however, come at the cost of communication overhead and sensitivity to network variability. Adaptive offloading strategies have been proposed to balance latency and accuracy under fluctuating conditions~\cite{chinchali2019network}. Foundational work in cloud robotics identifies key challenges, including latency, reliability, and transparent execution across heterogeneous resources~\cite{kehoe2015cloud}. More recent systems address these practical challenges. For example, FogROS2~\cite{ichnowski2023fogros2} integrates offloading into the Robot Operating System~2 (ROS2) framework and reports latency reductions of up to 50\% for SLAM and up to 28$\times$ speedups for motion planning using cloud GPUs.

\subsection{AI Model Acceleration for Edge Deployment}

Deploying deep learning models on embedded and edge platforms requires aggressive optimization to meet latency and power constraints. Vendor-specific inference engines have become the dominant solution for production deployments.
NVIDIA TensorRT~\cite{tensorrt2024} %is the de facto standard for GPU-accelerated inference,%
offers graph optimizations, layer fusion, and mixed-precision execution that achieve state-of-the-art latency on Jetson and data center NVIDIA GPUs. Intel OpenVINO~\cite{openvino2024} provides similar capabilities for Intel CPUs, integrated GPUs, and neural processing units, enabling efficient deployment on x86-based edge platforms. 
However, both solutions are tightly coupled to their respective hardware ecosystems and require their own runtime environments for execution.
ONNX Runtime~\cite{onnxruntime2023} offers a more portable alternative through interchangeable execution providers, but this flexibility comes at the cost of lower peak performance compared to vendor-specific stacks.

%However, these solutions share a common limitation: they produce optimized models that remain tied to their respective runtime environments. TensorRT requires the TensorRT runtime, OpenVINO requires the OpenVINO runtime, and ONNX Runtime requires its execution providers. Integrating any of these with an offloading framework demands a dedicated plugin—adding development overhead and introducing additional complexity. 

% Deep learning compilers address this limitation by generating optimized code from a common intermediate representation. TVM~\cite{chen2018tvm} produces optimized code for diverse hardware backends through automated search of optimization spaces, but its auto-tuning phase can require hours or days per model-hardware pair, limiting practical deployment in environments with evolving models or heterogeneous infrastructure. Moreover, TVM-produced shared libraries remain tied to the TVM runtime for execution. PyTorch 2.0 introduced \texttt{torch.compile}~\cite{ansel2024pytorch}, which applies graph capture and kernel fusion through the Inductor backend, achieving competitive inference performance with minimal user effort, yet remaining tightly coupled to the PyTorch runtime without producing portable standalone artifacts.

Deep learning compilers address this limitation by generating optimized code from a common intermediate representation. TVM~\cite{chen2018tvm} produces optimized code for diverse hardware backends through automated search of optimization spaces, but its auto-tuning phase can require hours or days per model-hardware pair, limiting practical deployment in environments with evolving models or heterogeneous infrastructure. Moreover, although TVM produces shared libraries, these remain tied to the TVM runtime for execution. PyTorch 2.0 introduced \texttt{torch.compile}~\cite{torchcompile}, which applies graph capture and kernel fusion through the Inductor backend to accelerate model execution. While \texttt{torch.compile} achieves competitive inference performance with minimal user effort, it remains tightly coupled to the PyTorch runtime and does not produce portable artifacts suitable for standalone deployment. 

The SOL compiler~\cite{weber2020sol} emerges as the only solution combining hardware portability with runtime independence, as it produces standalone shared libraries (.so) with no external runtime dependencies. These self-contained binaries expose well-defined C entry points that can be invoked directly via standard dynamic linking, eliminating the need for framework-specific plugins when integrating with offloading systems. Like TVM, SOL ingests models from PyTorch, TensorFlow, and ONNX and generates optimized binaries for diverse backends through mathematically equivalent transformations. However, SOL employs fast layer-wise auto-tuning that completes in minutes rather than hours, making it practical for robotic systems where models may need to be recompiled frequently or deployed across diverse hardware.

\subsection{Transparent Offloading Frameworks}
Beyond model optimization, recent research has explored frameworks that abstract heterogeneous acceleration resources to enable transparent remote execution. These approaches aim to decouple application logic from the underlying hardware and execution location, allowing inference tasks to run on local, edge, or cloud resources without modifying application code.
Several systems support remote acceleration. API remoting frameworks such as rCUDA~\cite{rcuda2010} and GVirtuS~\cite{gvirtus2010} intercept GPU API calls and forward them to remote servers with physical accelerators. While effective for CUDA workloads, these solutions are tightly coupled to specific low-level APIs and may incur significant communication overhead due to fine-grained call interception. Model-serving systems such as NVIDIA Triton~\cite{triton2024} provide remote inference over gRPC or HTTP, but require explicit client integration and do not offer execution transparency---applications must be modified to use the serving API.

vAccel~\cite{vaccel2024} addresses these limitations. Unlike rCUDA and GVirtuS, vAccel is hardware-agnostic: the same interface dispatches operations to CPUs, GPUs, or other accelerators without API-specific interception. Unlike Triton, vAccel does not require explicit client integration. Applications invoke operations through a unified GenOp interface, while the execution location is resolved transparently. By exposing coarse-grained operations instead of low-level API calls, it reduces communication overhead to a single round-trip per inference rather than per GPU call. Crucially, the GenOp interface directly accepts shared libraries with standard C entry points, enabling seamless integration with SOL-compiled binaries without any custom plugin or adaptation layer.

It is important to distinguish vAccel from robotics middleware such as ROS2~\cite{ros2}. ROS2 provides a distributed communication framework that enables software components (nodes) to exchange data across one or multiple machines, but it does not natively support transparent function-level computation offloading. Extensions such as FogROS2 \cite{ichnowski2023fogros2} enable the migration of ROS2 nodes to remote cloud or fog resources. However, this approach operates at node granularity and requires the ROS2 software stack to be deployed on the remote infrastructure. In contrast, vAccel operates at function granularity, offloading individual acceleratable operations rather than entire nodes. It therefore complements ROS2-based systems by offering lightweight, function-level offloading without requiring ROS2 middleware and communication stack to run on edge servers. In our framework, ROS2 manages sensor data distribution and communication between vision application components, while vAccel transparently offloads compute-intensive inference functions to edge resources.

% It is important to distinguish vAccel from robot middleware such as ROS2~\cite{ros2}. ROS2 provides a publish-subscribe communication layer based on DDS for coordinating distributed robot software, but does not natively support transparent computation offloading. While platforms such as FogROS2 extend ROS2 to enable cloud deployment of entire ROS nodes, this approach requires the full ROS stack on remote servers and operates at node granularity rather than function level. vAccel complements ROS-based systems by providing a lightweight, function-level offloading mechanism that integrates with robot applications without requiring ROS infrastructure on edge servers. In our framework, the robot application uses ROS2 for sensor data acquisition and inter-node communication, while vAccel handles the transparent offloading of compute-intensive inference operations to edge resources.
\section{The \name{} Framework}

\label{sec:architecture}

\begin{figure}[tb]
   \centering    
   \includegraphics[width=1\columnwidth]{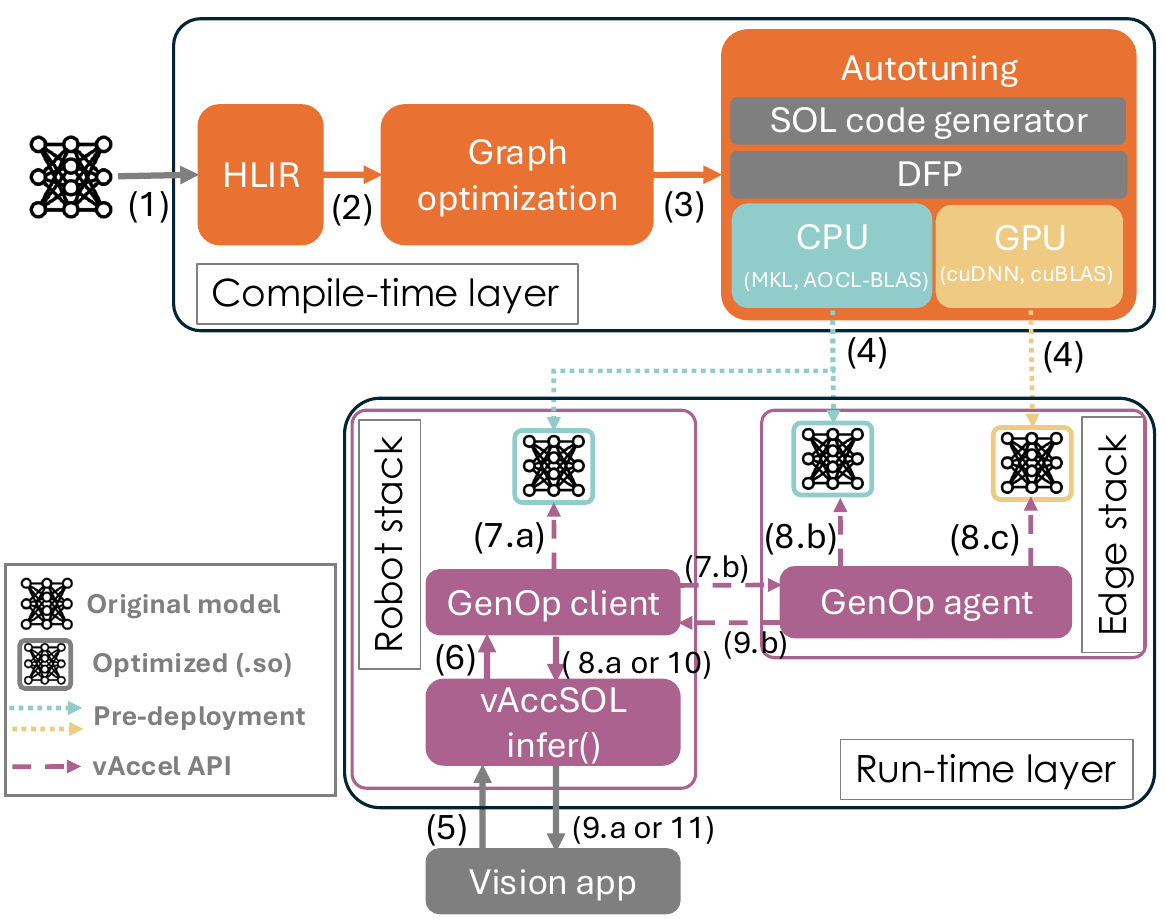}
      %\caption{Overview of the architectural building blocks.}
      \caption{\emph{\name{}} architecture}
      \vspace{-3mm}
      \label{fig:vAccSOL_architecture}
\end{figure}

\subsection{Overview}

Deploying deep learning vision models on mobile robots poses two coupled challenges: \textit{how} to execute inference efficiently on heterogeneous hardware, and \textit{where} to execute it under varying resource availability and network conditions. Existing solutions address these separately: inference engines optimize for specific hardware, while offloading frameworks focus on remote execution. \emph{\name{}} integrates both into a unified pipeline by combining SOL, a neural network compiler that generates hardware-optimized inference binaries, with vAccel, a runtime layer that transparently dispatches execution to local or remote resources.

Figure~\ref{fig:vAccSOL_architecture} illustrates the \emph{\name{}} architecture, organized into two layers. The \textit{compile-time layer} (top) transforms models into optimized shared libraries that are deployed on both robot and edge platforms. The \textit{run-time layer} (bottom) handles inference invocation and execution placement through vAccel's GenOp client-agent architecture. The vision application remains outside the framework boundary, interacting only through a single entry point: the \texttt{infer()} call.

The key property of \emph{\name{}} is execution transparency: the same application code, API calls, and model artifacts work across all execution modes. Switching between local and remote inference requires only a configuration change.

\subsection{Compile-Time Layer: Model Optimization with SOL}

The compile-time layer, shown in the upper portion of Figure~\ref{fig:vAccSOL_architecture}, transforms neural network models into optimized, hardware-specific executables. SOL begins by parsing the input model from supported frontends (e.g., PyTorch, TensorFlow, ONNX) and translating it into a proprietary High-Level Intermediate Representation (HLIR)~(step~1--2). This unified intermediate representation enables whole-graph compilation rather than interpretation of high-level scripts, thereby eliminating runtime overhead associated with frontend execution.

Once in HLIR form, SOL applies a series of global graph optimizations~(step~3), including dead code elimination, constant folding, and layer fusion. These transformations reduce redundant computation and improve data locality before any backend-specific lowering occurs.

The optimized graph is then partitioned into execution clusters and lowered to target-specific backends. SOL supports both its proprietary code generator, Depth-First Parallelization (DFP), and vendor libraries. On NVIDIA GPUs, convolution layers may be implemented via cuDNN or cuBLAS; on x86 CPUs, they may rely on MKL or AOCL-BLAS. Backend selection is performed through a combination of heuristics and autotuning, enabling heterogeneous backend mixing when beneficial. A key performance advantage arises from DFP code generation, which is particularly efficient for non-compute-intensive layers where general-purpose libraries may introduce unnecessary overhead~\cite{weber2018brainslug}.

The final output is a standalone shared library (\texttt{.so}) with minimal external dependencies. These SOL-generated artifacts are GenOp-compatible: they expose well-defined entry points that map directly to vAccel's generic operation interface, requiring no framework-specific plugins or custom integration logic. As shown in step~4, the same optimized binary is deployed on both robot and edge platforms, forming the computational backbone of the run-time layer.

\subsection{Run-Time Layer: Transparent Execution with vAccel}

The run-time layer, shown in the lower portion of Figure~\ref{fig:vAccSOL_architecture}, determines where inference is executed. vAccel provides a client-agent architecture that enables transparent invocation of compute-intensive operations across distributed resources.

The vision application invokes inference through a single entry point: the \texttt{infer()} call~(step~5). This call passes through the vAccel API~(step~6) to the GenOp client, which decides the execution path based on configuration. For local execution, the GenOp client loads and executes the SOL-compiled binary directly on the robot's CPU~(steps~7.a,~8.a). For remote execution, it forwards the request over the network to the GenOp agent deployed on the edge platform~(step~7.b).

The GenOp agent manages edge-side compute resources. Upon receiving a request ~(step~9.b), it loads the same SOL-compiled binary and dispatches execution to the appropriate backend—CPU or GPU—depending on availability~(steps~8.b,~8.c). Communication between the client and the agent uses a lightweight Remote Procedure Call (RPC) mechanism that transfers input tensors to the edge over a reliable transport (TCP or VSOCK). Results are then returned to the robot~(step~10 or~11), with inference round-trip latency primarily determined by the transfer of the model input and output tensors, as analyzed in Sec.~\ref{subsec:results-performance}.

From the application's perspective, inference is a single API call. Whether it executes on the robot's CPU or an edge GPU is determined by configuration, not code. This separation of concerns--SOL handles efficiency, vAccel handles placement--is the core design principle of \emph{\name{}}.

\section{Experimental Evaluation}
\label{sec:evaluation}

We validate \emph{\name{}} on a real-world testbed integrating a commercial quadruped robot and local edge infrastructure.

\subsection{Experimental Setup}
\label{subsec:testbed}

\begin{figure}[!t]
      \centering
      \includegraphics[width=1\columnwidth]{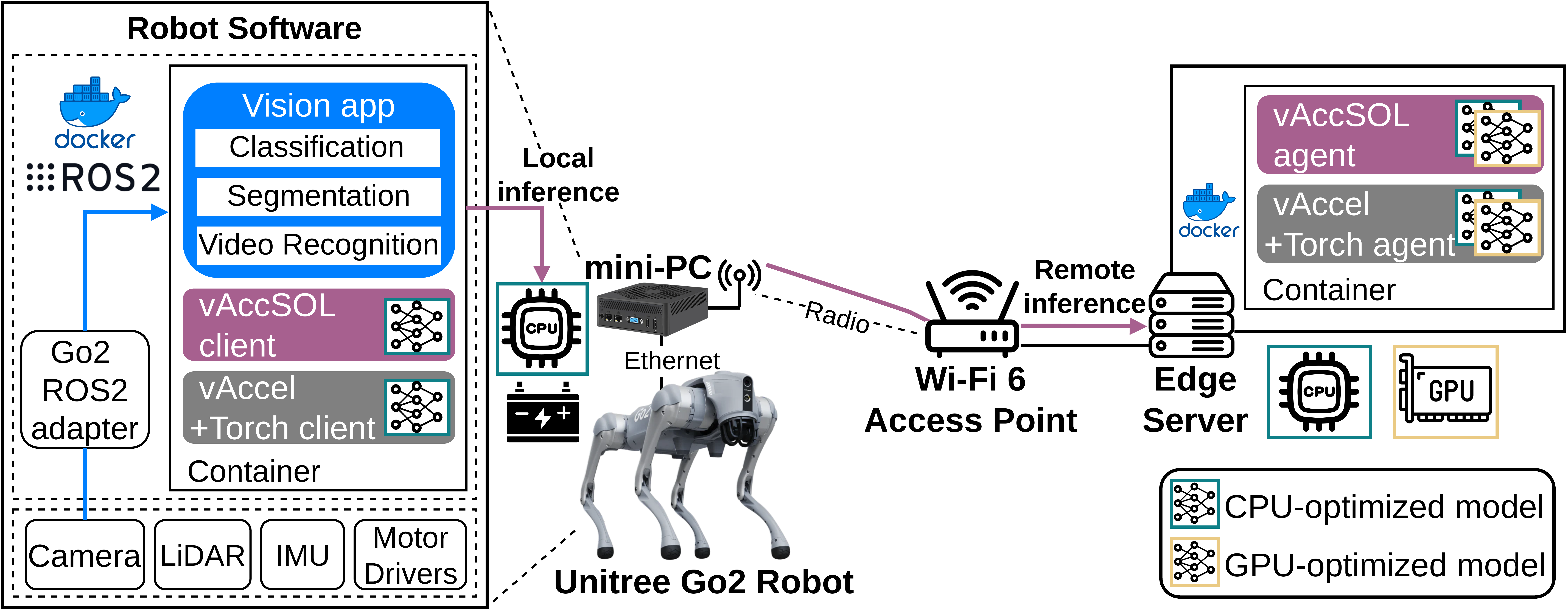}
      \caption{Experimental testbed.}
      \label{fig:testbed}
\end{figure}

Our experimental testbed (Fig.~\ref{fig:testbed}) consists of a Unitree Go2 EDU quadruped robot\footnote{\url{https://support.unitree.com/home/en/developer}}, a wireless access point, and a nearby edge server deployed in an indoor university hallway. The robot is equipped with a front-facing HD camera streaming video at 1280$\times$720 resolution and 15 frames per second (FPS) for real-time vision perception, along with an Inertial Measurement Unit (IMU) and a 4D LiDAR supporting SLAM. Since the robot does not provide a dedicated onboard computer for deploying custom AI applications, we attach an external mini-PC running Ubuntu~24.04 and powered by an onboard power bank. The mini-PC connects to the robot’s internal network via Gigabit Ethernet to access camera streams and executes the complete vision pipeline. It runs the robot-side \emph{\name{}} client, which intercepts inference calls from the vision application through the vAccel API. Depending on the execution configuration (Sec.~\ref{sec:architecture}), these calls are dispatched either to a locally loaded SOL-compiled shared library or to the remote edge agent with CPU/GPU resources.

The mini-PC is equipped with an Intel N100 CPU (4 cores, up to 3.4~GHz) and 12~GB RAM, and connects via a Wi-Fi~6 (AX200) interface through an access point (ASUS RT-AX89X) to the edge server. The edge server features an NVIDIA GeForce RTX~5080~Ti GPU, an Intel Core Ultra 9 275HX CPU (24 cores, up to 6.5~GHz), and 32~GB RAM. It hosts the \emph{\name{}} agent, which receives remote inference requests and executes them using SOL-optimized model artifacts. Importantly, switching between local and remote execution requires no modifications to the vision application running on the robot.

\begin{figure*}[!tb]
      \centering
      \includegraphics[width=0.95\textwidth]{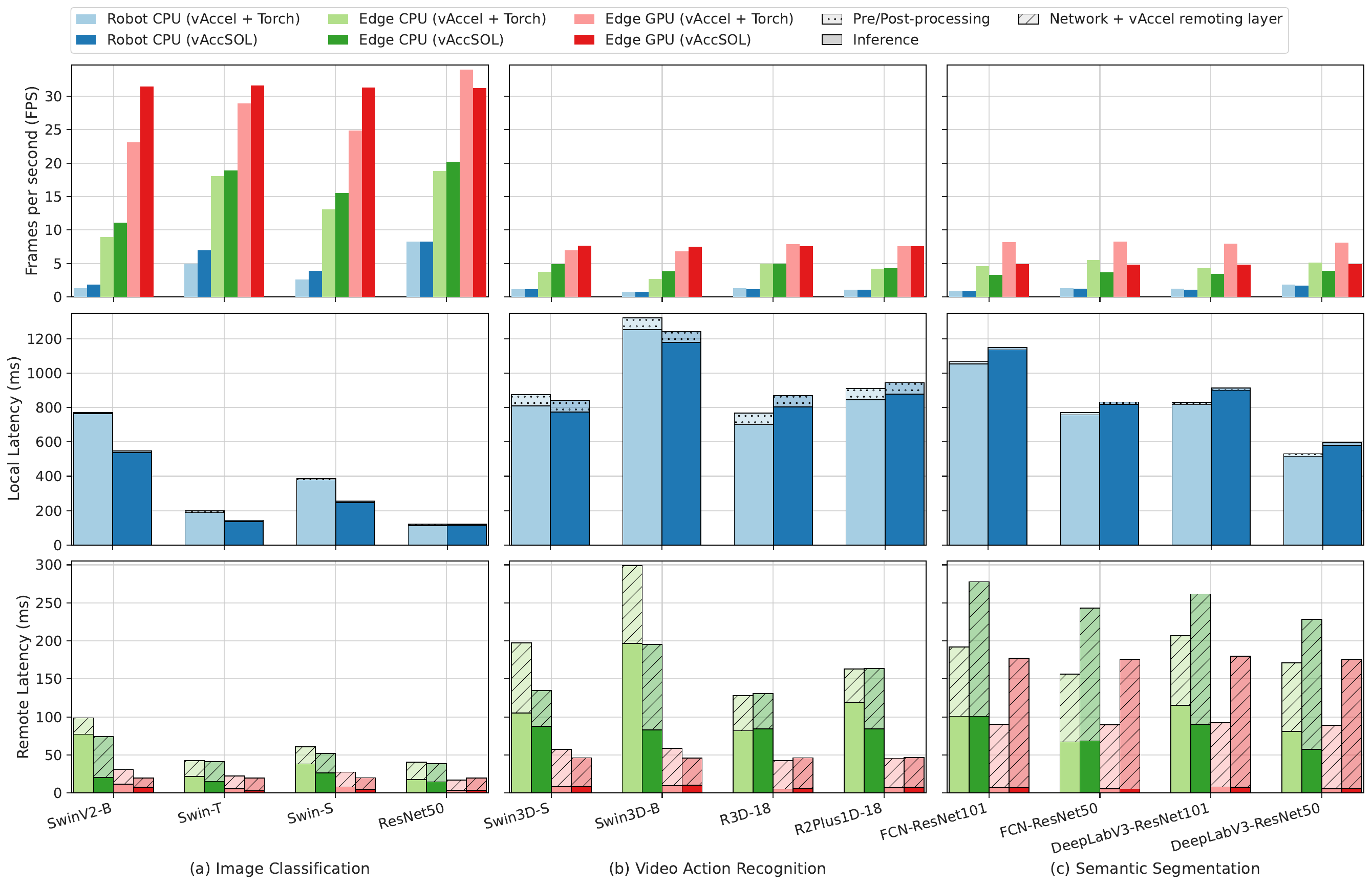}
    \caption{Frame rate and total processing latency of the robot vision application for single-camera input. Top: achieved frame rate (higher values indicate better performance). Middle and bottom: latency breakdown under local and edge execution (lower is better). We compare Torch and SOL under local execution and edge offloading. Results are grouped by task: (a) image classification, (b) video action recognition, and (c) semantic segmentation.}
      \label{fig:fps_and_latency_results}
      \vspace{-2mm}
\end{figure*}

\subsection{Robotic Vision Pipeline}
\label{subsec:pipeline}
The robot’s onboard sensors (camera, IMU, LiDAR) and actuators (motor drivers) run as native processes on its internal compute board. The vision application executes on the external mini-PC inside a Docker\footnote{\url{https://www.docker.com}} container using ROS2~\cite{ros2} and PyTorch\footnote{\url{https://pytorch.org}}. To access camera data in a standardized format compatible with other robotic applications, we deploy a ROS2 middleware adapter, which interfaces with the Go2 via the manufacturer’s SDK and exposes sensor data and control interfaces as ROS2 topics. During operation, the adapter receives the camera stream as H.264 video over UDP, decodes it, and republishes it as uncompressed image messages. The decoding step adds approximately 0.3~W of CPU power on the mini-PC, which is a negligible contribution compared to the total robot-side power reported in Sec.~\ref{subsec:results-power}.

The vision application subscribes to the image stream, performs preprocessing (frame resizing, normalization, tensor formatting), executes inference (locally or remotely), and applies task-specific post-processing to transform raw model outputs into application-level results (e.g., top-1 class selection or per-pixel segmentation masks). Results are then published to ROS2 topics for consumption by other robotic components. The application supports image classification, semantic segmentation, and video action recognition tasks.

\subsection{Evaluation Methodology}
\label{subsec:methodology}

We evaluate our solution in terms of (i) the achieved frame rate (FPS) of the robot-side vision application and (ii) CPU/GPU power consumption at both the robot and edge platforms.

\textbf{Performance Metrics.} End-to-end latency is measured across the complete vision pipeline, including image acquisition via the ROS2 adapter, pre-processing, model inference (local or remote), post-processing, and publication of results in ROS2. The system operates sequentially, processing one frame at a time without pipelining. For each frame, latency is recorded as the wall-clock time between image acquisition and result publication. Frame rate (FPS) is computed as the inverse of the average per-frame latency over all processed frames. CPU power on the robot mini-PC and the edge CPU is measured using the Intel RAPL interface~\cite{intel-rapl}, which provides hardware energy counters for processor domains. Average CPU power (W) is computed from the energy difference over time. Edge GPU power is monitored using the NVIDIA NVML API~\cite{nvml}, which reports real-time GPU power consumption. Power values are sampled during execution and averaged over each experiment.

\textbf{Models and Tasks.} We evaluate twelve representative deep learning models spanning convolutional, transformer-based, and hybrid spatiotemporal architectures in the three representative vision tasks that are considered relevant in mobile robotic operations: image classification (ResNet50, Swin-T, Swin-S, Swin-V2-B), video action recognition (R3D-18, R2Plus1D-18, Swin3D-S, Swin3D-B), and semantic segmentation (DeepLabV3-ResNet50/101, FCN-ResNet50/101). 

\textbf{Execution Configurations.} Each model is evaluated under two execution modes supported by \emph{\name{}}. In both modes, models are downloaded from the PyTorch repository and compiled offline before deployment (a process taking on the order of minutes). Since our focus is on runtime inference during robot operation, compilation time is excluded from the evaluation. We evaluate two backends: \emph{\name{}}, our primary framework based on the vAccel client remoting system with SOL-compiled binaries, and a standalone vAccel client using \texttt{torch.compile}~\cite{torchcompile} (Inductor backend), hereafter referred to as \emph{Torch}, which serves as the baseline state-of-the-art compiler. SOL-generated shared libraries are directly compatible with the GenOp interface, as discussed in Sec.~\ref{sec:architecture} and can therefore be invoked without additional integration layers or runtime dependencies. In contrast, the integration with Torch models required a dedicated software layer to manage model loading, tensor serialization, and execution through the libtorch runtime--introducing additional implementation complexity and a runtime dependency. In all experiments, Torch is used in its default configuration, with TF32 enabled and non-deterministic kernels allowed. This relaxes strict numerical reproducibility constraints and allows greater flexibility in kernel selection, fusion, and scheduling, ensuring a fair comparison between SOL and Torch.

\textit{Local Execution.} Models run on the robot CPU through their respective vAccel clients, which map inference calls to local resources. The \emph{\name{}} client loads and executes the SOL-compiled binary directly via the GenOp interface, while the standalone vAccel client manages execution through the libtorch runtime. 

\textit{Remote Edge Execution.} Models run remotely through their respective vAccel agents deployed at the network edge, enabling low-latency inference on edge CPU or GPU resources. The robot application invokes inference in the same manner as in local execution; the client forwards the request to the edge and returns the results to the robot. End-to-end latency includes network round-trip time for transmitting inputs and receiving outputs, and serialization/deserialization overhead of the vAccel remoting layer. 

Each experiment runs continuously for 60 seconds while the robot follows a pre-programmed trajectory at a constant speed of 0.5 m/s. To ensure fair comparison between SOL and Torch backends, all models use identical fixed-size inputs based on the task: (1, 3, 224, 224) for classification and segmentation, and (1, 3, 16, 112, 112) for video recognition, corresponding to (batch, channels, height, width) and (batch, channels, frames, height, width), respectively. We further verify that neither offloading nor backend selection affects model accuracy by evaluating the models on standard validation datasets with fixed inputs. Accuracy remains unchanged across execution modes and backends since identical model weights and input configurations are used. Detailed results are omitted for brevity.
\section{Evaluation Results}
\label{sec:results}

\subsection{Frame Rate and Latency Analysis}
\label{subsec:results-performance}

Figure~\ref{fig:fps_and_latency_results} presents the frame rate (FPS) and end-to-end latency measured at the robot application for single-camera processing. The upper subplots show the achieved FPS, while the lower subplots present the corresponding latency breakdown, organized by task: image classification (Fig.~\ref{fig:fps_and_latency_results}.a), video action recognition (Fig.~\ref{fig:fps_and_latency_results}.b), and semantic segmentation (Fig.~\ref{fig:fps_and_latency_results}.c). For each model, we compare inference using Torch (baseline) and SOL under local CPU execution and edge offloading to CPU or GPU resources.

Under local CPU execution on the robot, Torch and SOL achieve comparable performance across most models. SOL shows moderate improvements for transformer-based classifiers (e.g., Swin-S: 2.59 to 3.93~FPS; Swin-T: 5.02 to 6.97~FPS), while CNN-based models such as ResNet50 perform nearly identically under both backends. Conversely, Torch achieves slightly higher throughput for all semantic segmentation and some video recognition models (e.g., DeepLabV3-ResNet50: 1.88 vs.\ 1.68~FPS; R3D-18: 1.30 vs.\ 1.15~FPS). The latency breakdown in the middle plot further indicates that inference dominates end-to-end latency in all local configurations--typically above 95\% of the total--with pre- and post-processing contributing only a minor fraction. Performance differences thus reflect the distinct optimization strategies of each backend, with no consistent advantage observed across all architectures.

% This behavior aligns with the different optimization strategies of the two backends: Torch performs runtime compilation and device-specific autotuning on the target platform, whereas SOL executes a precompiled optimized binary.
% TORCH BETTER WITH DEFAULT CONFIG; SOL MORE DEPENDENT ON AUTO TUNING FOR BETTER PERFORMANCE

\begin{figure*}[!tb]
      \centering
      \includegraphics[width=0.95\textwidth]{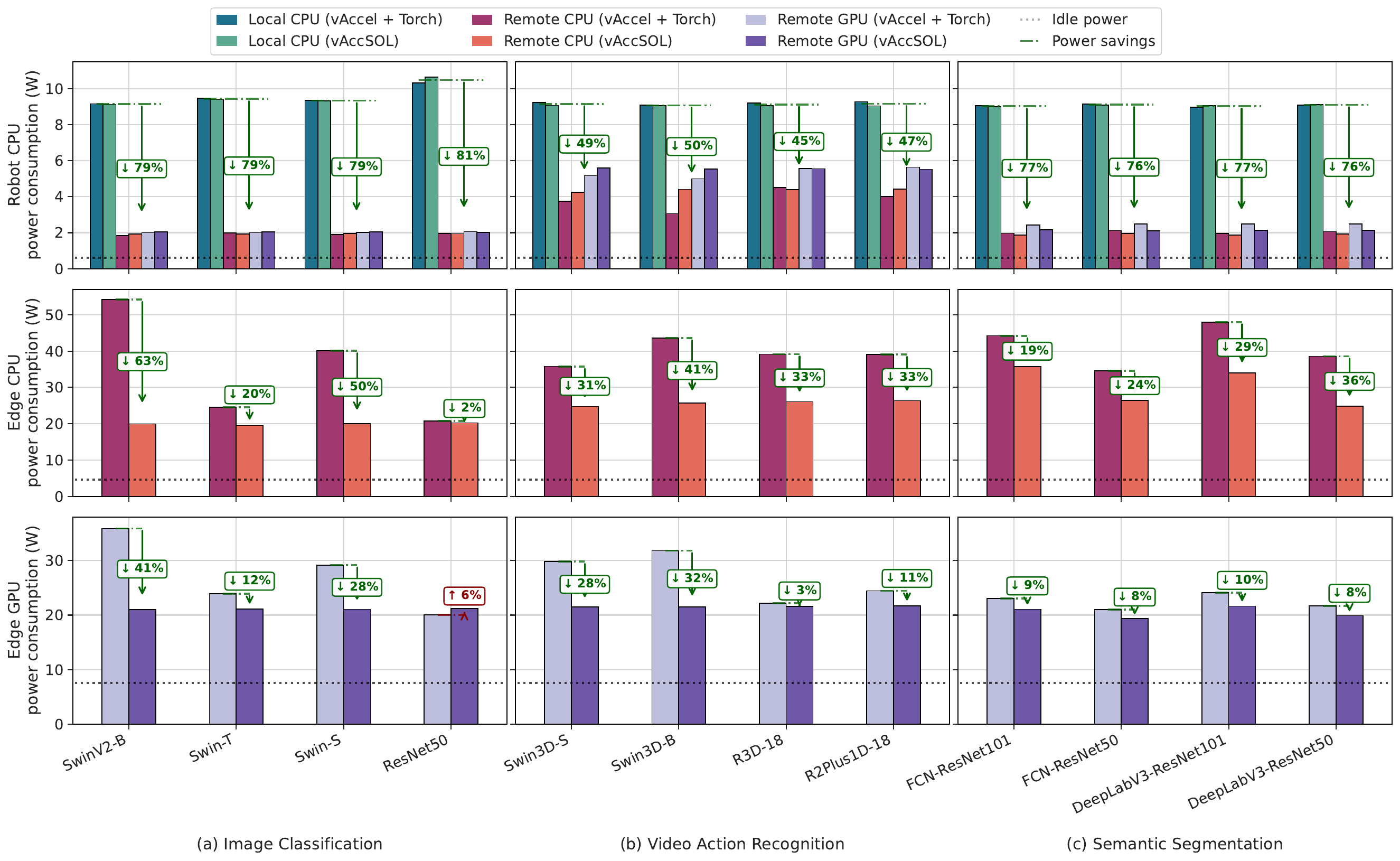}
      \caption{Power consumption of the robot CPU, edge CPU, and edge GPU during local and remote inference. Annotations quantify robot-side power savings from edge offloading (top) and edge-side power savings of SOL compared to Torch (middle, bottom).}
      \label{fig:power_consumption_results}
      \vspace{-2mm}
\end{figure*}

Offloading inference to the edge significantly increases throughput across all tasks and model families. Edge CPU execution provides approximately 2--6$\times$ higher throughput than robot-side execution. Edge GPU execution delivers the highest performance: image classification speedups reach 4--24$\times$ over local execution (e.g., ResNet50: 8.29 to 33.97~FPS; Swin-V2-B: 1.30 to 31.46~FPS), while video recognition achieves 6--9$\times$ (e.g., Swin3D-B: 0.80 to 7.53~FPS). For semantic segmentation, edge CPU offloading yields consistent 3--5$\times$ gains. Edge GPU throughput, however, is constrained by network transfer overhead, as analyzed below.

The latency breakdown explains the observed frame rate trends. Remote execution substantially reduces inference time thanks to the superior computational capability of the edge hardware. For example, Swin-V2-B decreases from 764~ms locally to 20~ms on the edge CPU and 7.6~ms on the edge GPU (38$\times$ and 100$\times$ reduction, respectively). As inference time decreases, network transfer and vAccel remoting overhead become the dominant cost in the end-to-end latency budget. This effect is most pronounced for tasks with large tensor payloads: semantic segmentation models incur $\sim$170~ms of network overhead due to large per-pixel output tensors, while video models face 35--80~ms driven by multi-frame input tensors--in both cases exceeding inference time by several times on the edge GPU.

Comparing SOL and Torch under remote execution, we observe that SOL achieves similar or lower inference latency across all edge configurations. Yet in several cases, particularly for segmentation models, higher network overhead offsets this advantage, yielding lower end-to-end frame rates on the robot side when executing on the edge GPU (e.g., FCN-ResNet50: 4.84 vs.\ 8.24~FPS with Torch).

We observe some inconsistencies in network overhead trends, which often—though not always—penalize \emph{\name{}} relative to vAccel with Torch. These differences stem from the distinct integration approaches: SOL relies on vAccel's generic GenOp interface, which minimizes integration effort but introduces modest serialization overhead, whereas Torch required a dedicated plugin with optimized tensor handling, achieving lower transport latency at the cost of significantly higher development effort.

Nevertheless, remote execution remains significantly faster than local execution for all models and both backends, confirming that the acceleration benefits of nearby edge infrastructure outweigh the network transfer costs even for the most bandwidth-demanding workloads.

\subsection{Power Consumption Analysis}
\label{subsec:results-power}

Figure~\ref{fig:power_consumption_results} reports power consumption during end-to-end inference for the robot CPU (top), edge CPU (middle), and edge GPU (bottom) across all evaluated models and execution configurations.

Under local execution, robot CPU power remains nearly constant at 9--10~W across all models and backends, reflecting sustained high CPU utilization. When inference is offloaded to the edge, robot power drops significantly--approaching the device idle baseline of 0.6~W in several configurations. For image classification and segmentation models, robot power decreases to approximately 1.8--2.5~W, representing a 76--80\% reduction compared to local execution. For video models, which involve higher data transfer and temporal buffering, robot power ranges from approximately 3--5.6~W, yielding a 45--51\% reduction. This behavior is consistent with the latency results (bottom subplots of Fig.~\ref{fig:fps_and_latency_results}): slower edge response allows the robot CPU to remain largely idle between frames, while faster GPU execution increases local activity slightly due to higher frame rates.

Edge power consumption reflects the computational load transferred from the robot. Under remote CPU execution, SOL reduces edge CPU power compared to Torch across most models--from modest savings for lightweight architectures (e.g., ResNet50: 20.8 vs.\ 20.3~W) to up to 63\% for heavier transformer models (e.g., Swin-V2-B: 54.2 vs.\ 20.0~W). This reduction aligns with the lower inference latency observed for SOL, indicating more efficient execution on the edge CPU. Under remote GPU execution, SOL consistently operates in the 19--22~W range across nearly all models, while Torch reaches up to 36~W for large transformer models (e.g., Swin-V2-B: 35.9~W). The exception is ResNet50, where both backends operate at comparable GPU power levels ($\sim$20--21~W). Despite the additional energy consumed at the edge, GPU acceleration significantly reduces robot-side power consumption while increasing frame rate, which is the primary focus of our analysis.

Examining edge CPU and GPU power consumption together, we observe that \emph{\name{}} consistently achieves lower power consumption than its Torch counterpart, with ResNet50 being the only exception. This advantage holds even in configurations where \emph{\name{}} exhibits higher network overhead due to faster inference: the energy cost of computation dominates that of data transfer, so reduced inference time translates directly to lower overall power consumption.

Overall, these results confirm that \emph{\name{}} effectively shifts computational load from the battery-constrained robot to stationary edge infrastructure. Although total system power increases, the robot experiences a substantial reduction in energy consumption, directly translating into longer operational endurance. This architectural design also offers a natural scalability advantage: since inference runs on shared edge infrastructure, multiple robots can be supported concurrently without increasing per-robot hardware cost or energy budget. The remaining constraints shift to edge capacity and network bandwidth--resources that are significantly easier to provision and upgrade than embedded robot hardware.

% Given that edge servers operate on fixed power  sources with far greater computational resources, this architectural choice also offers a natural scalability advantage: in multi-robot deployments, additional robots can share the same infrastructure without increasing per-robot hardware cost or energy budget, with scalability constrained only by edge server capacity and network bandwidth.
% It is important to note that our analysis prioritizes minimizing robot-side energy use, as onboard battery capacity ultimately limits deployment time.

% \section{Open challenges}
% Looking forward, several open challenges must be addressed to fully realize the potential of vision task offloading for quadruped robots in 6G networks. Wireless network variability demands adaptive offloading strategies that respond to fluctuating latency and bandwidth. Partial offloading, where tasks are split between onboard and edge execution, remains an open research problem, particularly for deep vision pipelines. Model heterogeneity across robots and edge nodes complicates deployment and learning, while privacy and security concerns arise from transmitting sensitive visual data. Emerging research directions point toward learning-driven offloading policies, federated and privacy-preserving inference, and digital twin–based optimization frameworks.

\section{Conclusions and future work}
\label{sec:conclusion}
This paper presented \emph{\name{}}, a unified framework for efficient and transparent execution of AI vision workloads on mobile robots. By combining compiler-driven model optimization with transparent computation offloading, \emph{\name{}} enables hardware-optimized inference to run either locally on the robot or remotely on nearby edge infrastructure without requiring modifications to the robot application. We validated the approach in a real-world testbed using a commercial quadruped robot and local edge infrastructure. Experimental results show that optimized local execution achieves comparable or improved performance over a Torch compiler baseline. With edge offloading, \emph{\name{}} reduces robot-side power consumption by up to 80\% compared to local execution and edge-side power consumption by up to 60\% compared to Torch, while increasing the vision pipeline frame rate by up to 24$\times$. These improvements enable mobile robots to support more demanding perception pipelines without compromising onboard energy constraints. During our evaluation in a controlled indoor environment, wireless communication with the edge remained stable and did not introduce observable performance degradation or execution failures.

Future work will extend the evaluation to additional robot and edge platforms to assess portability and hardware-dependent performance variations. We also plan to compare wireless technologies (e.g., Wi-Fi and 5G) under controlled variations in latency, bandwidth, and packet loss. Finally, the framework can be enhanced with adaptive offloading mechanisms that dynamically select the execution location based on real-time network conditions and robot energy consumption.

%\section*{ACKNOWLEDGMENT}

\bibliographystyle{IEEEtran} 
\bibliography{references} 

\end{document}